\documentclass[journal]{IEEEtran}

\usepackage[T1]{fontenc}
\usepackage{amsmath,amssymb}
\usepackage{graphicx,booktabs,xcolor}
\definecolor{fillblue}{RGB}{42,85,121}

\usepackage{cite}
\usepackage{url}
\usepackage[hidelinks]{hyperref}
\hypersetup{
  pdftitle={Lyapunov-Inspired Sustained Balance on Dynamically Responsive Supports},
  pdfauthor={Changyan Li, Guiyin Tian, Jing Xia, Shiyan Hu},
  pdfsubject={LISST: Lyapunov-Inspired Sustained Stabilization},
  pdfkeywords={balance control, coupled dynamics, legged robots, Lyapunov methods, physics-informed learning, robot-environment interaction}
}

\begin{document}

\title{Lyapunov-Inspired Sustained Balance\\
on Dynamically Responsive Supports}
\author{Changyan Li\textsuperscript{1}, Guiyin Tian\textsuperscript{1}, Jing Xia\textsuperscript{1}, and Shiyan Hu\textsuperscript{1}\thanks{\textsuperscript{1}Department of Data and Systems Engineering, The University of Hong Kong, Hong Kong SAR, China.}}
\markboth{}{}
\maketitle

\begin{abstract}
Conventional balance formulations often assume a rigid, stationary support, decoupling body stabilization from support motion. This assumption becomes inadequate on mechanically responsive supports, creating a coupled control challenge: corrective body motions intended to restore equilibrium can also excite support motion. We formulate sustained balance as a coupled regulation problem that coordinates equilibrium restoration with suppression of motion induced by the robot's own corrections. We introduce LISST (Lyapunov-Inspired Sustained Stabilization), a learning framework that accounts for how earlier actions shape the conditions for subsequent corrections. It uses recent observations to support immediate adjustments and longer histories to characterize persistent motion, with physical supervision relating the learned representation to robot--support behavior. To evaluate corrective actions, LISST uses coupled dynamics to assess both imbalance and the direction of motion. A Lyapunov-inspired criterion measures progress over finite windows, accommodating the transients required to restore balance. This physical evaluation guides joint-level policy learning through backpropagation through differentiable model rollouts. Privileged guidance focuses training on difficult states reached by the deployment policy and is transferred through distillation. The resulting policy operates using only onboard proprioception. Real-world experiments demonstrate that LISST maintains balance while limiting residual oscillations during continuous interaction with a dynamically responsive support. Video: \href{https://youtu.be/iGHU7S4wPyw}{\textcolor{fillblue}{youtu.be/iGHU7S4wPyw}}.
\end{abstract}


\section{Introduction}
\label{sec:introduction}

\IEEEPARstart{B}{alancing} on a mechanically responsive support requires corrective movements that also alter the supporting surface. Whole-body controllers coordinate body motion and contact forces~\cite{sentis2010compliant,righetti2013optimal}, which also affect the support~\cite{nava2017modeling}. Shifting the robot can restore static moment balance, while the acceleration and braking needed to reach that configuration can excite support motion. The resulting robot--support motion can persist after the static imbalance has been compensated.

Each subsequent correction therefore begins with the motion left by earlier actions. We study \emph{sustained balance} as the continuous regulation of coupled robot--support motion to establish and maintain equilibrium. We investigate this problem with a quadruped on a passive, pitch-hinged board under changing endpoint loads, preserving physical state and controller history throughout the interaction. The robot chooses body and contact adjustments without a prescribed reference motion, using onboard proprioception at deployment.

Output-feedback controllers stabilize simplified robots on unstable supports without direct environmental feedback~\cite{nagarajan2014balancing}, while coupled robot--support models describe and control their shared dynamics~\cite{nava2017modeling}. For quadrupeds on moving platforms, LAS-MP combines adaptive postural control with proprioceptive robot and platform state estimation~\cite{yoon2024lasmp}. In sustained balance, each correction begins with the motion left by earlier actions, which can change whether the next movement suppresses or amplifies the response. A proprioceptive policy therefore needs information about this ongoing response and an action evaluation that accounts for it.

We introduce LISST (Lyapunov-Inspired Sustained Stabilization), a learning framework that uses interaction history to inform corrective actions. Similar body postures can occur during quiet balance or at an oscillation turning point, so posture alone does not identify the ongoing response. History-based locomotion and adaptation use past observations to infer control context~\cite{lee2020learning,kumar2021rma}. For sustained balance, longer histories characterize persistent motion, while recent observations support immediate adjustments.

Interpreting motion also requires a criterion for whether a correction advances balance. Equal relative kinetic energy can correspond to motion toward or away from equilibrium, and the acceleration required for correction acts back on the support. LISST uses coupled dynamics to evaluate these directional effects and a Lyapunov-inspired criterion to assess progress over finite windows, allowing for corrective transients. This physical evaluation guides the joint-target policy through backpropagation through time (BPTT) in differentiable world-model rollouts~\cite{fussell2021supertrack,hafner2020dreamer}, connecting the simplified dynamics to the actual action interface.

Training must also account for the difficult states created by the deployment policy's own corrections. LISST directs privileged guidance to these states, preserving their accumulated motion and observation history, and transfers it through context distillation. The deployed policy requires only proprioception.

Our main contributions are as follows:
\begin{itemize}
\item \textbf{Structured Temporal Representation-Based Control.}
We develop a policy that uses short-history context and supervised joint/body estimates for immediate adjustments, together with a structured temporal representation of ongoing motion. Learned signals are parameterized by offset, amplitude, frequency, and phase and related to robot--support motion through physical reconstruction and short-term continuation. These complementary inputs let the Actor condition corrective actions on both the current body state and the evolving motion pattern.
\item \textbf{Physics-Informed Lyapunov-Inspired Optimization.}
Coupled dynamics guide policy optimization by evaluating imbalance together with the direction of relative motion. A load-aware finite-window Lyapunov-inspired decay criterion accommodates corrective transients, and BPTT through differentiable world-model rollouts carries this physical evaluation to the joint-target policy. The analysis relates realized window decrease to conditional convergence and boundedness.
\item \textbf{Residual-Guided Privileged Policy Learning.}
Bounded residual corrections to the deployment context provide privileged guidance through a shared Actor. Some privileged rollouts start from low-value and pre-failure states encountered by the deployment policy, preserving physical state, load context, observation history, and reward memory. Distillation on states collected by the deployment policy transfers this guidance to the situations arising from the policy's own interaction with the support.
\end{itemize}

\section{Related Work}
\label{sec:related}

\subsection{Dynamic Balance and Robot--Environment Interaction}
Model-based balance control coordinates whole-body motion with contact forces and support constraints. Sentis \emph{et al.}~\cite{sentis2010compliant} regulate center-of-mass behavior and internal forces, while Righetti \emph{et al.}~\cite{righetti2013optimal} optimize contact-force distribution through inverse dynamics. Capturability analysis relates stopping without falling to foot placement, support geometry, and angular momentum~\cite{koolen2012capturability}. Trajectory optimization and model predictive control account for contact interactions over finite horizons~\cite{posa2014direct,dicarlo2018dynamic}.

Moving supports introduce additional dynamics. Lee and Goswami~\cite{lee2012momentum} regulate momentum on nonstationary ground, while Gao \emph{et al.}~\cite{gao2025swaying} use time-varying foot-placement control for walking on swaying surfaces. Nagarajan and Yamane~\cite{nagarajan2014balancing} stabilize simplified planar robots on unstable supports without direct environmental feedback. Nava \emph{et al.}~\cite{nava2017modeling} model robot--seesaw dynamics and contact constraints and demonstrate coupled stabilization in simulation. LAS-MP~\cite{yoon2024lasmp} learns adaptive postural control with proprioceptive robot and platform state estimation. These studies motivate treating the support as part of the dynamics. Our focus is the continuing response to corrective actions, including the motion that remains after static compensation.

\subsection{Learning-Based Legged Locomotion and Adaptation}
Reinforcement learning enables quadrupeds to acquire motor skills in simulation and transfer them to physical robots~\cite{hwangbo2019learning}. The information available to the policy shapes its ability to adapt: Lee \emph{et al.}~\cite{lee2020learning} use proprioceptive history for locomotion on challenging terrain, whereas Miki \emph{et al.}~\cite{miki2022learning} combine proprioceptive and exteroceptive feedback. RMA~\cite{kumar2021rma} infers an adaptation context from recent observations and actions, and Margolis \emph{et al.}~\cite{margolis2024rapid} combine a command curriculum with online system identification.

Privileged learning transfers simulation information to policies with restricted deployment inputs~\cite{lee2020learning,kumar2021rma}. The states on which guidance is provided also matter. DAgger~\cite{ross2011dagger} addresses the expert--learner state-distribution mismatch through supervision on learner-visited states.

Beyond the information and guidance available to a policy, learning depends on how its actions are evaluated. Learned world models predict action-conditioned transitions for planning or policy improvement through imagined trajectories~\cite{hafner2019latent,hafner2025worldmodels}. Physical structure can guide how those trajectories are evaluated. CLF-RL~\cite{li2026clfrl} constructs tracking and Lyapunov-decay rewards from model-generated reference motions, and finite-step control Lyapunov methods assess decrease over multiple transitions~\cite{noroozi2019finitestep}. Our objective addresses load compensation without a prescribed reference motion. It uses the physical model to evaluate signed corrective motion jointly with imbalance and support motion, retaining the reaction of acceleration on the support.

\subsection{State Estimation and Motion Representation}
State estimation recovers feedback variables from sensor measurements. Bloesch \emph{et al.}~\cite{bloesch2012state} fuse leg kinematics and inertial measurements, while Hartley \emph{et al.}~\cite{hartley2020contact} use contact-aided invariant filtering to estimate pose, velocity, and contact points. Camurri \emph{et al.}~\cite{camurri2017probabilistic} estimate contact reliability and detect impacts. On moving ground, He \emph{et al.}~\cite{he2026noninertial} estimate robot state relative to noninertial ground using encoders and IMUs on both the robot and support.

Temporal encoders and adaptation modules extract control context from interaction history~\cite{lee2020learning,kumar2021rma}, while DreamWaQ~\cite{nahrendra2023dreamwaq} learns implicit terrain information from proprioception. DeepPhase~\cite{starke2022deepphase} learns periodic embeddings of full-body motion, and Fourier Latent Dynamics~\cite{li2024fld} uses a predictive Fourier representation for motion encoding and generation. For sustained balance, the representation must describe the motion already in progress when the next correction begins. This motivates relating latent temporal coordinates to robot--support motion through physical supervision.

We next formulate the coupled mechanics and the proprioceptive control interface on which LISST operates.

\section{Problem Formulation}
\label{sec:problem}

\subsection{Task and State Description}
We consider a quadruped on a rigid board that rotates passively about a fixed pitch hinge at $p^H$. The robot shifts its body and relocates its feet to compensate changing endpoint loads, without a prescribed reference motion or balancing location. Load updates preserve the mechanical state and controller history throughout each continuous episode.

For a fixed load with feasible balance, the target is a level, stationary board, zero net static torque, and vanishing robot motion relative to the board while maintaining support. The robot configuration satisfying these conditions can vary with the load.

In a right-handed inertial frame, $\mathbf e_z$ points upward and $\mathbf e_y$ is the hinge axis. The board is level at $\theta=0$, and positive pitch follows the right-hand rule about $\mathbf e_y$. Downward loads of magnitudes $F_+(t)$ and $F_-(t)$ act at opposite board locations with local longitudinal coordinates $+\ell$ and $-\ell$.

The mechanical state is
\begin{equation}
 s_t=(p_t,R_t,v_t,\omega_t,q_t,\dot q_t,\theta_t,\dot\theta_t),
 \label{eq:mechanical_state}
\end{equation}
where $p,R,v,\omega$ describe the robot base position, orientation, linear velocity, and angular velocity in the inertial frame, and $q\in\mathbb R^{12}$ contains joint angles. Physical parameters and external-load context are denoted by $\xi_t$. Mechanical parameters are held fixed within an episode.

\subsection{Coupled Robot--Board Dynamics}
Body and foot adjustments change the contact forces coupling robot and board motion. Let $q_r=(p,R,q)$ and $\nu_r=(v,\omega,\dot q)$ denote the robot configuration and generalized velocity. We adapt the coupled robot--environment formulation in~\cite{nava2017modeling} to a fixed-hinge board with ideal point contacts. During continuous contact, the dynamics are
\begin{equation}
 \begin{aligned}
 M_r\dot\nu_r+h_r &= S^{\mathsf T}\tau+J_r^{\mathsf T}\lambda+w_r,\\
 I_B\ddot\theta+b_B\dot\theta
     &=\tau_L-J_B^{\mathsf T}\lambda+w_B.
 \end{aligned}
 \label{eq:mechanics}
\end{equation}
Here $M_r$ is the robot inertia matrix, $h_r$ collects gravity, velocity-dependent, and passive robot terms, $S$ selects the actuated joints, and $\tau$ denotes motor torque. The board has hinge inertia $I_B$ and viscous damping $b_B$. Its centered mass distribution gives zero self-weight moment about the hinge. The signed load moment is $\tau_L=\ell\cos\theta(F_+-F_-)$. Other external forces and contact reactions are collected in $w_r$ and $w_B$, including hinge-stop and board--ground reactions where present.

The vector $\lambda$ stacks foot--board contact forces acting on the robot, expressed in the inertial frame. $J_r$ maps robot velocity to contact-point velocity. At contact $i$, with position $p_i$, the board-side Jacobian is $J_{B,i}=\mathbf e_y\times(p_i-p^H)$. Vertically stacking these contact Jacobians gives $J_B$. The opposite signs of the contact terms in~\eqref{eq:mechanics} follow from action and reaction. For active sticking contacts, equality of robot and board contact-point velocities yields
\begin{equation}
 \begin{aligned}
 J_r\nu_r &= J_B\dot\theta,\\
 J_r\dot\nu_r-J_B\ddot\theta
     &=-\dot J_r\nu_r+\dot J_B\dot\theta.
 \end{aligned}
 \label{eq:contact_coupling}
\end{equation}
Together with~\eqref{eq:mechanics}, the differentiated constraint couples robot acceleration, board acceleration, and contact force. These equations describe active sticking contacts, with admissible forces restricted by unilateral contact and friction. Slipping, lift-off, and landing change the active constraints.

Let $H_{r,H,y}$ denote robot angular momentum about the fixed hinge $p^H$, projected onto $\mathbf e_y$. The robot contact moment $J_B^{\mathsf T}\lambda$ cancels the opposite board moment, giving the aggregate balance
\begin{equation}
 \begin{aligned}
 \frac{d}{dt}\big(H_{r,H,y}+I_B\dot\theta\big)
     &=\tau_{\mathrm{net}}-b_B\dot\theta+\tau_{\mathrm{other}},\\
 \tau_{\mathrm{net}}
     &=g\sum_j m_j x_j+\ell\cos\theta(F_+-F_-),
 \end{aligned}
 \label{eq:coupled_hinge_momentum}
\end{equation}
where $x_j=\mathbf e_x^{\mathsf T}(p^{C_j}-p^H)$ is the world-horizontal position of link $j$'s center of mass relative to the hinge, $m_j$ is its mass, and $\tau_{\mathrm{other}}$ collects other external moments. Away from stops and additional contacts, static equilibrium requires $\tau_{\mathrm{net}}=0$. Mass redistribution changes this moment to compensate the load. Angular momentum and relative robot--board motion can persist after static compensation, so both imbalance and motion enter the objective.

\subsection{Proprioceptive Control Problem}
The deployed controller must regulate this coupled motion using onboard proprioception. It receives a proprioceptive sensor frame,
\begin{equation}
 o_t=\big[\widetilde{q_t-q^0},\;\widetilde{\dot q_t},\;
       \widetilde{R_t^{\mathsf T}\omega_t},\;
       \widetilde{g_t^b},\;a_{t-1}\big],
 \label{eq:proprioceptive_observation}
\end{equation}
where $q^0$ is the default joint pose, $g_t^b=-R_t^{\mathsf T}\mathbf e_z$ is projected gravity, and tildes denote sensor noise and latency. The previous action is the issued normalized command. Board state, loads, and physical parameters are excluded from deployment inputs. Sensor history supplies context for the response to earlier loads and actions.

A policy maps observation history $\mathcal H_t=(o_{t-L+1},\ldots,o_t)$ to joint commands $a_t=\pi_\vartheta(\mathcal H_t)$. The deployment interface requests
\begin{equation}
 q_t^{\mathrm{target}}=q^0+\alpha\odot a_t,
 \label{eq:joint_targets}
\end{equation}
where $\alpha$ contains joint action scales. Low-level servos and actuator limits determine the resulting torques $\tau$.

The policy must maintain foot--board support, compensate loads, and suppress residual motion under contact, joint, and actuation constraints. Simulator state supplies training objectives and supervision, while deployment uses~\eqref{eq:proprioceptive_observation}--\eqref{eq:joint_targets}.

\section{Method}
\label{sec:method}

\begin{figure*}[t]
\centering
\includegraphics[width=\textwidth]{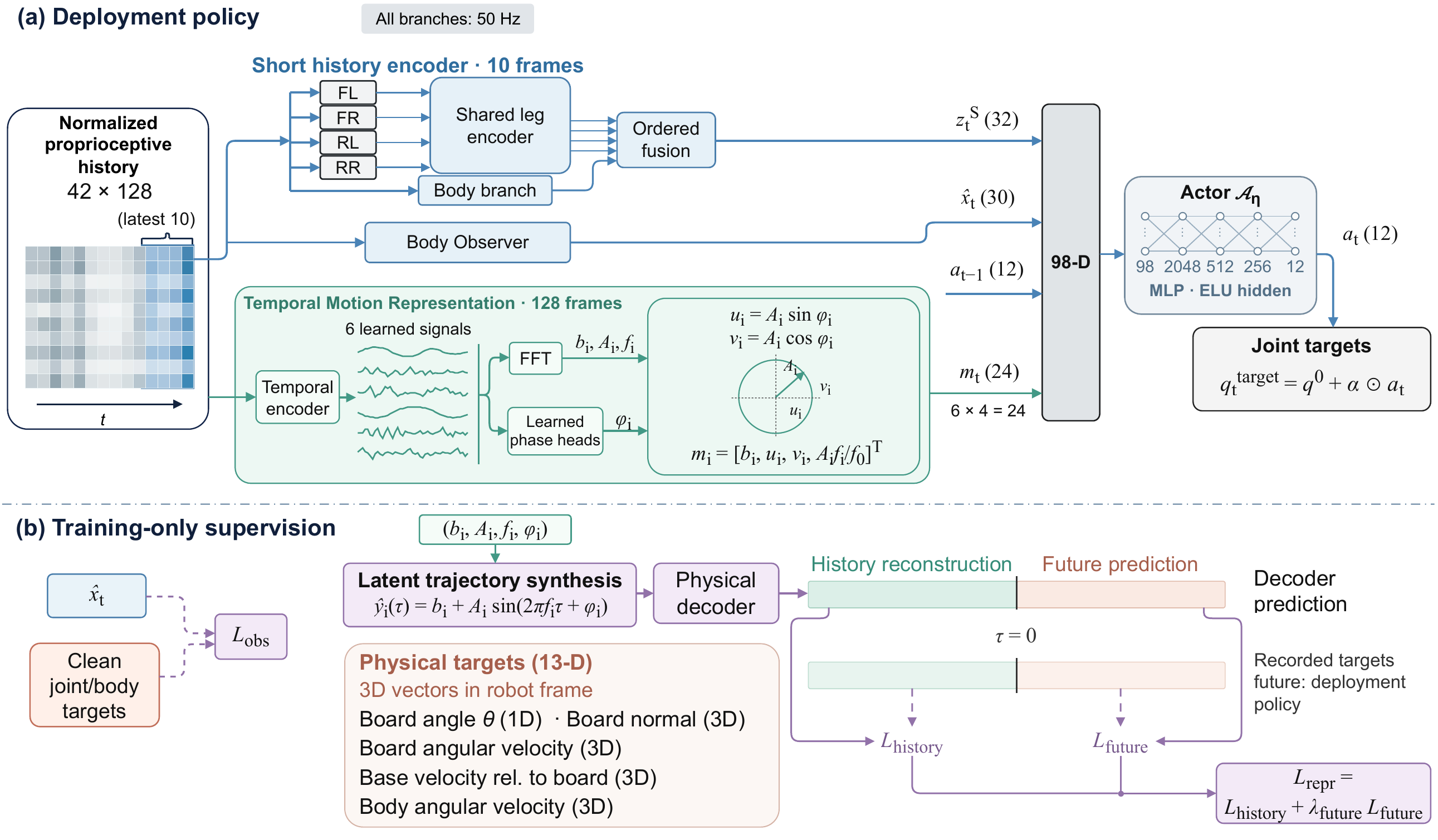}
\caption{Structured temporal representation-based control in LISST. (a) Short-history context, joint/body estimates, and longer-history motion features enter the Actor with the previous command. All branches share an update rate. (b) Clean targets supervise the Body Observer, while a physical decoder reconstructs history and short-term continuation from latent trajectories. Future targets follow recorded deployment-policy behavior. The decoder provides training supervision, not Actor inputs. The separate world model supplies action-conditioned rollouts. Waveforms and history tiles are schematic.}
\label{fig:proprioceptive_feedback}
\end{figure*}

\subsection{Overview}
\label{sec:method_overview}
LISST connects a history-dependent policy to a dynamics-informed evaluation of its actions. The deployment loop estimates the ongoing response from proprioception, while training uses physical state to supervise the representation and evaluate corrective progress. We use $S$ and $T$ to denote the deployment and privileged policies, respectively.

The control loop in Fig.~\ref{fig:proprioceptive_feedback} draws on both recent observations and a longer history. Short-history context and joint/body estimates support immediate adjustments. The longer history describes how motion has evolved beyond the current posture through parameterized latent signals, which physical reconstruction and short-term continuation relate to robot--support motion. The Actor maps these inputs to joint targets, and the resulting response becomes part of the history for its next decision. All branches update at the same control frequency, although their history lengths differ.

The training objective incorporates physical priors about compensation and motion suppression. In the reduced dynamics, relative velocity changes the compensating moment, and acceleration reacts on the support. These couplings determine quadratic weights that evaluate motion direction jointly with static imbalance and board motion. Adding the remaining whole-body relative energy gives a common measure of balance, complemented by auxiliary terms for posture, support, and execution requirements. A soft Lyapunov decay criterion then evaluates progress over finite windows: it accommodates the transients needed for correction and accounts separately for external-load changes. Under the stated feasibility and realized-decrease conditions, the analysis establishes window-end decay or ultimate bounds. During training, the world model predicts the consequences of joint targets, and the resulting rewards guide the Actor through backpropagation through time. Physical information thus shapes both what the representation describes, through reconstruction supervision, and how corrective actions are evaluated, through the dynamics-derived objective.

Privileged learning concentrates guidance on difficult states reached by the deployment policy (Fig.~\ref{fig:resilience_privileged_learning}). Some privileged rollouts begin from its low-value and pre-failure states, carrying forward physical state, load context, sensor history, and reward memory. A bounded residual refines the deployment context, and the shared Actor maps both roles' contexts to actions. Distillation transfers this guidance on states collected by the deployment policy.

\subsection{Structured Temporal Representation-Based Control}
\label{sec:multiscale_feedback}
The representation separates immediate context and measurement correction from the temporal structure of persistent motion (Fig.~\ref{fig:proprioceptive_feedback}). Short-history branches supply the former, while the longer-history branch parameterizes learned signals by offset, amplitude, frequency, and phase.

\subsubsection{Short-History Encoder and Body Observer}
The short-history encoder learns task-dependent control context from recent joint deviations, joint velocities, and previous commands. A shared leg network processes these inputs, with leg identities and ordered fusion preserving the four-leg arrangement. Fusion with recent body gyro and projected gravity gives context $z_t^S$.

The Body Observer uses the same history for a different purpose: correcting noisy current measurements. Its output $\widehat x_t$ contains normalized joint deviations, joint velocities, body gyro, and projected gravity, supervised by clean joint/body variables through residual correction, with both architectures and their supervision specified in Appendix~\ref{app:feedback_implementation}.

\subsubsection{Temporal Motion Representation}
Recent observations support immediate adjustments, while a longer segment provides context for the magnitude and timing of persistent motion. The longer-history branch makes this temporal structure explicit through the sinusoidal latent parameterization introduced in DeepPhase and adopted by FLD~\cite{starke2022deepphase,li2024fld}. Its amplitude and phase coordinates describe the size of learned oscillatory signals and the current position within their local cycles, so the Actor can condition a correction on the motion already in progress.

A temporal encoder maps sensor history to learned signals, from which the fast Fourier transform (FFT) extracts frequency, amplitude, and offset. Channel-wise phase heads locate the current sample within each signal, giving the reconstructed latent trajectory
\begin{equation}
 \widehat y_i(\tau)=b_i+A_i\sin(2\pi f_i\tau+\phi_i),
 \label{eq:local_fourier_trajectory}
\end{equation}
Here $\tau=0$ is the current sample, and negative times index history. The offset $b_i$ describes the baseline, $A_i$ and $f_i$ describe local oscillation, and $\phi_i$ aligns it at $\tau=0$. Recomputing these parameters from the sliding history at every control step lets the local description change as the response evolves.

The Actor receives continuous amplitude-weighted phase coordinates $u_i=A_i\sin\phi_i$ and $v_i=A_i\cos\phi_i$. The channel-wise offsets, phase coordinates, and amplitude-weighted frequencies form the motion feedback vector:
\begin{equation}
 \begin{aligned}
 m_i&=\left[b_i,\;u_i,\;v_i,\;A_i f_i/f_0\right]^{\mathsf T},\\
 m_t&=[m_1^{\mathsf T},\ldots,m_6^{\mathsf T}]^{\mathsf T}
       \in\mathbb R^{24},
 \end{aligned}
 \label{eq:fourier_feedback_features}
\end{equation}
The fixed scale $f_0$ normalizes frequency. Amplitude weighting attenuates the phase and frequency contributions of inactive channels while retaining their offsets. The motion feedback enters the Actor alongside the short-history context.

Physical supervision gives these latent coordinates a connection to balance. The training-only decoder in Fig.~\ref{fig:proprioceptive_feedback}(b) maps reconstructed latent trajectories to the board hinge angle relative to level and four robot-frame vectors: board normal, board angular velocity, base velocity relative to the board, and body angular velocity. History reconstruction supervises observed motion, while short-term continuation supervises its local evolution under recorded behavior. The two losses are normalized separately to form $L_{\mathrm{repr}}$ (Appendix~\ref{app:feedback_implementation}). The physical decoder predicts continuation under recorded behavior without taking a candidate future action sequence as input, whereas the world model predicts action-conditioned transitions for policy optimization.

For the deployment policy, the shared Actor fuses the feedback branches and previous command:
\begin{equation}
 a_t=\mathcal A_\eta\big([z_t^S,\;\widehat x_t,\;a_{t-1},\;m_t]\big).
 \label{eq:student_feedback_actor}
\end{equation}
Its output specifies joint offsets through~\eqref{eq:joint_targets}. Deployment retains the short-history encoder, Body Observer, temporal encoder, FFT, phase heads, and Actor.

\subsection{Physics-Informed Lyapunov-Inspired Optimization}
\label{sec:energy_objective}
The moment balance in~\eqref{eq:coupled_hinge_momentum} provides the physical basis for the objective. A local model relates signed relative velocity to compensation and acceleration reaction, yielding a quadratic state cost. We add the remaining whole-body kinetic energy and use finite-window decay to evaluate corrective progress in the training reward.

\subsubsection{Coupled State Cost}
Compensation depends on how the robot moves along the board, so the state cost includes signed longitudinal motion relative to the support. Let $M=\sum_{j=1}^{13}m_j$ be the actual robot mass and let $\mathbf t_B=(\cos\theta,0,-\sin\theta)^{\mathsf T}$ be the board's longitudinal unit vector. With the board rigid velocity field $v^B(p)=\omega^B\times(p-p^H)$ and $\omega^B=\dot\theta\mathbf e_y$, define
\begin{equation}
 \begin{aligned}
 v_j^{\mathrm{rel}}&=v^{C_j}-v^B(p^{C_j}),\qquad
 \omega_j^{\mathrm{rel}}=\omega_j-\omega^B,\\
 v_{\parallel}&=\mathbf t_B^{\mathsf T}
                  \frac{1}{M}\sum_j m_jv_j^{\mathrm{rel}}.
 \end{aligned}
 \label{eq:longitudinal_relative_velocity}
\end{equation}
The signed longitudinal center-of-mass velocity $v_{\parallel}$ is measured relative to the board's rigid velocity field. The training objective computes it from simulator state and uses the normalized state
\begin{equation}
 \bar e=\left[\frac{\theta}{\theta_0},\;
       \frac{\dot\theta}{\Omega_0},\;
       \frac{\tau_{\mathrm{net}}}{Mg d_0},\;
       \frac{v_{\parallel}}{v_0}\right]^{\mathsf T}.
 \label{eq:scaled_balance_error}
\end{equation}

A constant-height point-mass reduction makes the acceleration reaction explicit. For longitudinal position $x$, relative speed $v_{\parallel}=\dot x$, nominal mass $M_0$, and height $h_0$, the hinge momentum is $[I_{B,0}+M_0(x^2+h_0^2)]\dot\theta+M_0h_0v_{\parallel}$. Near a level, fixed-load equilibrium, it gives
\begin{equation}
 \begin{aligned}
 \dot\theta&=\omega_B,\\
 I_0\dot\omega_B&=\tau_n-b_0\omega_B-M_0h_0a_{\parallel},\\
 \dot\tau_n&=M_0g v_{\parallel}+M_0gh_0\omega_B,\\
 \dot v_{\parallel}&=a_{\parallel}.
 \end{aligned}
 \label{eq:local_reaction_model}
\end{equation}
Here $\omega_B=\dot\theta$, $\tau_n$ is the reduced model's net static torque, and $a_{\parallel}$ is a virtual relative acceleration. Position changes compensate the load, while the acceleration that changes relative velocity also acts on the board through $-M_0h_0a_{\parallel}$. A linear-quadratic regulator (LQR) trades state deviation against acceleration effort in this model, and its quadratic value function supplies the fixed weight $P_4\succ0$ (derived in Appendix~\ref{app:nominal_quadratic}).

The point-mass reduction omits other whole-body motion that can persist during correction. To include this motion, we start from the full relative kinetic energy,
\begin{equation}
 \begin{aligned}
 K_{\mathrm{rel}}={}&\frac12\sum_j\left[
       m_j\|v_j^{\mathrm{rel}}\|^2+
       (\omega_j^{\mathrm{rel}})^{\mathsf T}I_j^W\omega_j^{\mathrm{rel}}
       \right]\\
   &+\frac12\sum_k J_k^{\mathrm{arm}}\dot q_k^2.
 \end{aligned}
 \label{eq:relative_kinetic_energy}
\end{equation}
Because longitudinal center-of-mass motion already enters the quadratic state cost, we subtract its kinetic-energy contribution and retain the remainder:
\begin{equation}
 K_{\mathrm{rem}}=K_{\mathrm{rel}}-\tfrac12 Mv_{\parallel}^2\geq0.
 \label{eq:remaining_relative_energy}
\end{equation}
The mass-weighted definition in~\eqref{eq:longitudinal_relative_velocity} makes this remainder nonnegative, as shown by the sum-of-squares form in Appendix~\ref{app:objective_implementation}. Adding it to the quadratic gives the coupled cost
\begin{equation}
 V(s,\xi)=\bar e^{\mathsf T}P_4\bar e+
        \frac{K_{\mathrm{rem}}}{\frac12 M v_0^2}.
 \label{eq:energy_cost}
\end{equation}
At the same board state and static imbalance, opposite longitudinal velocities can have equal kinetic energy. The quadratic cross terms distinguish how each velocity aligns with the remaining imbalance and board motion, while the energy remainder retains other translational, rotational, and joint motion. Through its LQR-derived weight $P_4$~\cite{tedrake2023underactuated}, the cost captures the nominal coupling between compensation and acceleration reaction. Trajectory rewards then guide the Actor toward joint targets in~\eqref{eq:joint_targets} that produce favorable corrections.

\begin{figure*}[t]
\centering
\includegraphics[width=\textwidth]{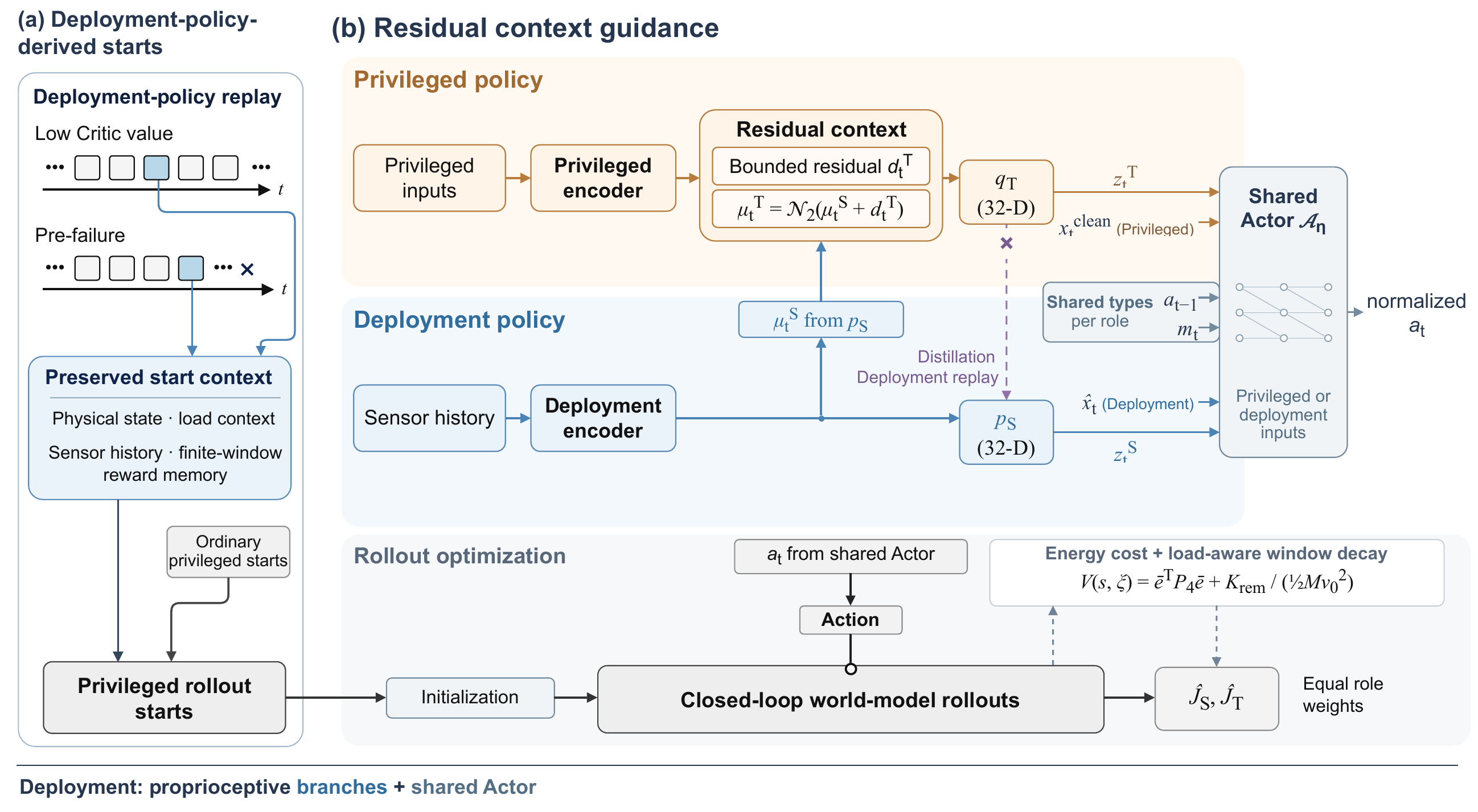}
\caption{Residual-guided privileged policy learning. (a) Low-value and pre-failure states collected by the deployment policy replace part of the privileged rollout starts, preserving physical state, load context, sensor history, and reward memory. (b) Bounded context residuals guide a shared Actor with role-specific inputs. Distillation stops gradients through the privileged distribution. World-model rollouts optimize equally weighted role returns using the reward in Section~\ref{sec:energy_objective}. The ordinary deployment start sampler is omitted. Auxiliary supervision appears in Fig.~\ref{fig:proprioceptive_feedback}(b). Deployment retains the proprioceptive branches and shared Actor.}
\label{fig:resilience_privileged_learning}
\end{figure*}

\subsubsection{Lyapunov Decay and Load-Aware Relaxation}
A Lyapunov function assigns a nonnegative scalar to departure from an equilibrium or equilibrium set, with zero on the target. If this value decreases along closed-loop trajectories, its evolution can establish stability and convergence under suitable conditions. We use the coupled cost in this spirit to encourage reduction of imbalance and relative motion. For fixed load and smooth closed-loop evolution, a sufficient condition for exponential decrease~\cite{tedrake2023underactuated} is
\begin{equation}
 \begin{aligned}
 \dot V&\leq-\kappa V,\qquad\kappa>0,\\
 \Longrightarrow\quad V(t)&\leq e^{-\kappa t}V(0).
 \end{aligned}
 \label{eq:ideal_lyapunov_decay}
\end{equation}
The cost describes the present balance condition, while its decrease describes progress. Corrective motion can temporarily raise the cost while redistributing mass. We therefore assess progress relative to a fixed window origin, following a finite-step Lyapunov viewpoint~\cite{noroozi2019finitestep}, and use a soft penalty for departures from the decay criterion. A fixed reference permits cost increases between adjacent steps while the trajectory remains below the reference decay envelope.

An external-load update can increase $V$ before the robot moves. The budget $B_t$ accumulates these positive increments between the reference and the current evaluation, separating a change in the task from the cost of the robot's correction. Motion-induced increases remain part of that correction cost.

We divide each continuous trajectory into consecutive windows of at most $N$ transitions. The reference remains fixed within each window and is renewed every $N$ transitions. For reference cost $V_{\mathrm{ref}}$ and $1\leq h\leq N$ elapsed transitions of duration $\Delta$, we define the load-aware soft decay criterion
\begin{equation}
 \begin{aligned}
 V_t&\leq e^{-\kappa h\Delta}V_{\mathrm{ref}}
             +B_t+\epsilon h/N+\zeta_t,\\
 \zeta_t&\geq0.
 \end{aligned}
 \label{eq:soft_decay_violation}
\end{equation}
Here $\epsilon\geq0$ sets a tolerance scaled by the available window fraction $h/N$. The smallest nonnegative slack $\zeta_t$ measures the remaining violation. We evaluate the criterion after every transition within the window and penalize the normalized slack $\delta_t=\zeta_t/(1+V_{\mathrm{ref}})$. Together, the fixed reference, soft violation penalty, and load budget distinguish corrective progress from external changes. Their parameters and memory updates are specified in Appendix~\ref{app:objective_implementation}.

The reward evaluates current balance quality and recent progress separately, with auxiliary terms for physical requirements:
\begin{equation}
 r_t=
 \underbrace{r_t^{\mathrm{bal}}}_{\text{balance benefit}}
 -\underbrace{\ell_t^{\mathrm{dec}}}_{\text{decay penalty}}
 +\underbrace{r_t^{\mathrm{phys}}}_{\text{auxiliary terms}}.
 \label{eq:learning_reward}
\end{equation}
The balance benefit is $r_t^{\mathrm{bal}}=6\Delta w_{\mathrm{sup},t}/(1+V_t)$, where $\Delta$ is the control-step duration. The smooth support weight $w_{\mathrm{sup},t}\in[0,1]$ depends on foot--board proximity, board footprint, and base clearance. At a fixed support weight, lower coupled cost yields greater benefit.

Contact transitions during foot relocation remain part of the evaluated trajectory. Any resulting violation of the decay criterion contributes to the slack penalty, while the support weight modulates the balance benefit.

The decay penalty is $\ell_t^{\mathrm{dec}}=\Delta H_+(\delta_t)$, with one-sided Huber function $H_+$. It increases with the normalized violation $\delta_t$ of~\eqref{eq:soft_decay_violation} and remains active independently of the support weight. The auxiliary reward $r_t^{\mathrm{phys}}$ regularizes body uprightness and height, contact geometry, joint and motor limits, action variation, and motor effort (Appendix~\ref{app:objective_implementation}).

Policy optimization uses closed-loop rollouts of an action-conditioned world model learned from training trajectories. Joint targets drive predicted transitions, which update the feedback history for subsequent decisions. The reward evaluates each trajectory through the energy cost, load-aware decay, and auxiliary terms. Backpropagation through time (BPTT) carries return gradients through reward evaluation, state transitions, and the joint-target mapping to the Actor and its policy encoders~\cite{fussell2021supertrack,hafner2020dreamer}. The dynamics-derived metric remains fixed: it determines how outcomes are evaluated, while world-model derivatives connect those outcomes to joint commands. World-model and Critic parameters are held fixed during these policy updates, with input derivatives retained.

On board-supported trajectories with feasible zero-cost states, fixed loads and zero tolerance and slack give exponential decay at non-overlapping window endpoints. Bounded load-induced cost increments and normalized violations below the contraction margin yield ultimate bounds on $\bar e$ and the full $K_{\mathrm{rel}}$. Appendix~\ref{app:objective_implementation} derives these bounds under the stated conditions on realized closed-loop decrease.

\subsection{Residual-Guided Privileged Policy Learning}
\label{sec:privileged_learning}
Privileged learning must connect additional physical information to the context used by the deployment policy. We express that guidance as a bounded residual around the deployment context mean, making the correction relative to its current representation. The shared Actor interprets both roles' contexts through the same action mapping. Rollout-start selection determines where this guidance is learned, and context distillation transfers it on states collected by the deployment policy.

The short-history encoder supplies the deployment context mean $\mu_t^S$. The privileged residual branch corrects this mean using current proprioception, clean body state, board-relative geometry, true base velocity, and contact information (Fig.~\ref{fig:resilience_privileged_learning}(b)):
\begin{equation}
 \begin{aligned}
 d_t^T&=c_T\tanh(\bar d_t^T),\\
 \mu_t^T&=\mathcal N_2(\mu_t^S+d_t^T),
 \end{aligned}
 \label{eq:privileged_residual_mean}
\end{equation}
where $\bar d_t^T$ is an encoder output, $c_T>0$ bounds each correction coordinate, and $\mathcal N_2$ denotes unit-vector normalization. Both roles use the shared Actor $\mathcal A_\eta$ and context sampling rule. The privileged policy substitutes clean body input for the Body Observer estimate, while both roles compute motion feedback from their sensor histories.

Part of the ordinary privileged rollout starts is replaced by low-value and nonterminal pre-failure states from the deployment policy (Fig.~\ref{fig:resilience_privileged_learning}(a)). Preserving physical state, load context, sensor history, and finite-window reward memory lets these rollouts continue from the original feedback and reward context. The privileged policy learns corrective actions from these deployment-policy states.

On states collected by the deployment policy, $L_{\mathrm{KL}}$ supervises its pre-projection Gaussian context with the privileged distribution as the target. With gradients stopped through the entire privileged distribution, this supervision transfers the residual guidance to the short-history encoder used at deployment.

Using the rollout optimization in Section~\ref{sec:energy_objective}, we compute imagined returns $\widehat J_T$ and $\widehat J_S$ for the privileged and deployment policies. Their joint objective combines equally weighted returns with context distillation and physical supervision:
\begin{equation}
 \begin{aligned}
 L_\pi={}&-\tfrac12\left(\mathbb E[\widehat J_T]+
                              \mathbb E[\widehat J_S]\right)\\
        &+\beta_{\mathrm{aux}}\left(L_{\mathrm{KL}}+
             L_{\mathrm{obs}}+\lambda_{\mathrm{repr}} L_{\mathrm{repr}}\right).
 \end{aligned}
 \label{eq:joint_policy_objective}
\end{equation}
Body Observer parameters receive only reconstruction gradients, although return differentiation retains derivatives with respect to its inputs. Return gradients pass through the Actor to the temporal encoder and phase heads. Privileged-policy returns also reach the deployment policy's short-history encoder through the residual mean. The losses $L_{\mathrm{obs}}$ and $L_{\mathrm{repr}}$ supervise the Body Observer and temporal motion representation, respectively. The latter uses the physical decoder and~\eqref{eq:representation_objective}, with sampling and gradient paths detailed in Appendix~\ref{app:training_implementation}.

Deployment follows~\eqref{eq:student_feedback_actor}, retaining the proprioceptive branches and shared Actor.

\section{Experiments}
\label{sec:experiments}
We evaluate LISST on a Unitree Go2 Edu quadruped on a passive, pitch-hinged board. The board measures $2.0\,\mathrm{m}\times0.45\,\mathrm{m}$ and is mounted at a height of $0.25\,\mathrm{m}$. Successive changes in endpoint loads elicit corrective robot--support interaction. The policy issues joint targets from onboard proprioception, without board-state or external-load measurements. Physical state and controller history persist across load changes, so each correction begins with the motion left by earlier actions.

Board angle and angular velocity characterize equilibrium deviation and residual board motion, alongside maintenance of support. These quantities assess whether corrections restore balance while limiting the motion carried into subsequent interactions. In real-world deployment, LISST maintains balance while limiting residual oscillations during continuous interaction with the support. The accompanying \href{https://youtu.be/iGHU7S4wPyw}{hardware video} presents the demonstration.

\section{Conclusion}
\label{sec:conclusion}
LISST treats sustained balance on dynamically responsive supports as joint regulation of equilibrium error and coupled motion. Temporal representations characterize the ongoing response, while a physics-informed, Lyapunov-inspired criterion guides optimization through differentiable rollouts. Privileged guidance focuses training on difficult states encountered by the deployment policy.

Real-world deployment demonstrates balance maintenance with limited residual oscillations during continuous robot--support interaction. This interaction-centered formulation evaluates corrections by both their immediate effect on equilibrium and the motion inherited by subsequent actions.

\bibliographystyle{IEEEtran-doi}
\bibliography{references}

\clearpage
\appendices
\section{Coupled Dynamics and Energy Construction}
\label{app:nominal_quadratic}

\subsection{Reduction from Hinge Momentum}
Consider a nominal point mass $M_0$ at constant board-relative height $h_0$ and longitudinal coordinate $x$. Its world coordinates relative to the fixed hinge are $X=x\cos\theta+h_0\sin\theta$ and $Z=-x\sin\theta+h_0\cos\theta$. With $v_{\parallel}=\dot x$, its hinge angular momentum is
\begin{equation}
 H_{r,H,y}=M_0(Z\dot X-X\dot Z)
          =M_0(x^2+h_0^2)\omega_B+M_0h_0v_{\parallel}.
 \label{eq:reduced_hinge_momentum}
\end{equation}
The robot--board contact forces cancel in the aggregate moment balance. Omitting other external moments, differentiation gives
\begin{equation}
 \begin{aligned}
 I(x)\dot\omega_B
   &=\tau_n-b_0\omega_B
       -2M_0xv_{\parallel}\omega_B-M_0h_0a_{\parallel},\\
 I(x)&=I_{B,0}+M_0(x^2+h_0^2),\\
 \tau_n&=M_0g(x\cos\theta+h_0\sin\theta)+\tau_L(\theta).
 \end{aligned}
 \label{eq:reduced_nonlinear_dynamics}
\end{equation}
For constant endpoint loads, $\tau_L(\theta)=\tau_{L,0}\cos\theta$. A level equilibrium satisfies $x^\star=-\tau_{L,0}/(M_0g)$, $v_{\parallel}=\omega_B=0$, and $\tau_n=0$. Linearization gives~\eqref{eq:local_reaction_model}, with $I_0=I_{B,0}+M_0[(x^\star)^2+h_0^2]$. Differentiating the static torque yields $\dot\tau_n=M_0g v_{\parallel}+M_0gh_0\omega_B$ to first order. Load jumps enter the objective budget separately.

This reduction assumes small deviations about level balance, constant height, and no additional contact or hinge-stop moments. Whole-body shape changes and actuator dynamics are omitted.

\subsection{Normalized Model and Quadratic Weight}
For the weighting calculation, use~\eqref{eq:scaled_balance_error} with $M_0,\tau_n$ in place of the actual mass and static torque, and let $u=a_{\parallel}/a_0$. Then
\begin{equation}
 \begin{aligned}
 \dot{\bar e}&=A_n\bar e+B_nu,\\
 A_n&=\begin{bmatrix}
 0&\Omega_0/\theta_0&0&0\\
 0&-b_0/I_0&M_0gd_0/(I_0\Omega_0)&0\\
 0&h_0\Omega_0/d_0&0&v_0/d_0\\
 0&0&0&0
 \end{bmatrix},\\
 B_n&=\begin{bmatrix}
 0&-M_0h_0a_0/(I_0\Omega_0)&0&a_0/v_0
 \end{bmatrix}^{\mathsf T}.
 \end{aligned}
 \label{eq:nominal_balance_model}
\end{equation}
The virtual acceleration input couples changes in relative velocity and static torque to the board reaction.

For the state and input weights in Table~\ref{tab:objective_parameters}, consider the normalized nominal cost $\int_0^\infty(\bar e^{\mathsf T}Q\bar e+r_uu^2)\,dt$. The input weight sets the nominal tradeoff between state regulation and relative acceleration. The positive stabilizing Riccati solution $\bar P_4$ satisfies
\begin{equation}
 A_n^{\mathsf T}\bar P_4+\bar P_4A_n-
 \bar P_4B_nr_u^{-1}B_n^{\mathsf T}\bar P_4+Q=0.
 \label{eq:nominal_riccati}
\end{equation}
Normalize $P_4=\bar P_4/\bar P_{4,11}$ to set its first diagonal entry to one.

\subsection{Relation to the Joint-Target Policy}
Under the nominal feedback $u^\star=-r_u^{-1}B_n^{\mathsf T}\bar P_4\bar e$, the quadratic obeys
\begin{equation}
 \frac{d}{dt}(\bar e^{\mathsf T}P_4\bar e)
 =-\frac{\bar e^{\mathsf T}Q\bar e+r_u(u^\star)^2}{\bar P_{4,11}}.
 \label{eq:nominal_only_decrease}
\end{equation}
Joint targets produce an effective relative acceleration through the servos and contact dynamics. Writing the local reduced motion as $\dot{\bar e}=A_n\bar e+B_nu_{\mathrm{eff}}+d$, where $d$ collects residual dynamics, adds $2\bar e^{\mathsf T}P_4[B_n(u_{\mathrm{eff}}-u^\star)+d]$ to~\eqref{eq:nominal_only_decrease}. This term separates the effect of realized motion from nominal quadratic decrease.

Policy learning uses $P_4$ to evaluate state quality. The finite-window criterion evaluates the full cost, including $K_{\mathrm{rem}}$, along closed-loop trajectories.

\subsection{Numerical Configuration}
\label{app:objective_parameters}
Table~\ref{tab:objective_parameters} lists the state scales and nominal design parameters, with angles converted to radians for computation. Episode-specific physical parameters determine masses, center-of-mass offsets, inertias, and armatures. The actual total mass sets the torque and kinetic-energy scales, whereas the nominal model and quadratic weight remain fixed across load updates and randomized episodes.

\begin{table}[!ht]
\caption{Objective scales and nominal model parameters.}
\label{tab:objective_parameters}
\centering\small
\begin{tabular}{@{}ll@{}}
\toprule
Parameter & Value \\
\midrule
\multicolumn{2}{@{}l}{\textit{State and input scales}}\\
$\theta_0$, $\Omega_0$ & $4^\circ$, $10^\circ/\mathrm{s}$\\
$d_0$, $v_0$ & $0.1\,\mathrm m$, $0.2\,\mathrm{m/s}$\\
$a_0$ & $1\,\mathrm{m/s^2}$\\
\midrule
\multicolumn{2}{@{}l}{\textit{Nominal dynamics and cost}}\\
$h_0$, $x^\star$ & $0.30\,\mathrm m$, $0$\\
$g$ & $9.81\,\mathrm{m/s^2}$\\
$b_0/I_0$ & $0.5\,\mathrm{s^{-1}}$\\
$M_0gd_0/(I_0\Omega_0)$ & $20\,\mathrm{s^{-1}}$\\
$Q$, $r_u$ & $\operatorname{diag}(1,1,0.1,1)$, $1$\\
\midrule
\multicolumn{2}{@{}l}{\textit{Implied inertia ratios}}\\
$I_0/M_0$ & $0.281036\,\mathrm{m^2}$\\
$I_{B,0}/M_0$ & $0.191036\,\mathrm{m^2}$\\
\bottomrule
\end{tabular}
\end{table}

These choices give the normalized matrices
\begin{equation}
 \begin{aligned}
 A_n&\simeq\begin{bmatrix}
 0&2.5&0&0\\
 0&-0.5&20&0\\
 0&0.523599&0&2\\
 0&0&0&0
 \end{bmatrix},\\
 B_n&\simeq[0,\;-6.116208,\;0,\;5]^{\mathsf T}.
 \end{aligned}
 \label{eq:numerical_nominal_model}
\end{equation}
For these matrices, the normalized Riccati weight is
\begin{equation}
 P_4\simeq\begin{bmatrix}
 1.000000&1.186256&4.783405&1.676711\\
 1.186256&2.868492&12.470522&4.391151\\
 4.783405&12.470522&56.682415&19.999484\\
 1.676711&4.391151&19.999484&7.284800
 \end{bmatrix}.
 \label{eq:nominal_P}
\end{equation}
The calculation uses the full-precision solution. The displayed matrices and implied inertia ratios are rounded.

\section{Reward Evaluation and Conditional Convergence}
\label{app:objective_implementation}

\subsection{Relative Kinetic-Energy Decomposition}
The identity $\sum_jm_j\mathbf t_B^{\mathsf T}v_j^{\mathrm{rel}}=Mv_{\parallel}$ gives
\begin{equation}
 \begin{aligned}
 K_{\mathrm{rem}}={}&\frac12\sum_jm_j
             \|v_j^{\mathrm{rel}}-v_{\parallel}\mathbf t_B\|^2\\
 &+\frac12\sum_j(\omega_j^{\mathrm{rel}})^{\mathsf T}
               I_j^W\omega_j^{\mathrm{rel}}
   +\frac12\sum_kJ_k^{\mathrm{arm}}\dot q_k^2.
 \end{aligned}
 \label{eq:energy_remainder_squares}
\end{equation}
The sum-of-squares form evaluates~\eqref{eq:remaining_relative_energy} without cancellation. The board velocity field is affine, so its mass-weighted value equals the field at the center of mass. Subtracting the collective longitudinal translation therefore leaves the transverse, relative link, and rotational motion.

\subsection{Load-Induced Cost Increments}
At a fixed mechanical state $s_k$, let $\xi_k^-$ and $\xi_k^+$ denote load contexts before and after an update. Endpoint-load changes leave mass, velocities, and $P_4$ fixed, changing only the third coordinate of $\bar e$. Hence
\begin{equation}
 \begin{aligned}
 \Delta\bar e_k&=\left[0,\;0,\;
        \frac{\Delta\tau_{L,k}}{Mg d_0},\;0\right]^{\mathsf T},\\
 \Delta V_k^L&=2(\bar e_k^-)^{\mathsf T}P_4\Delta\bar e_k+
                 \Delta\bar e_k^{\mathsf T}P_4\Delta\bar e_k.
 \end{aligned}
 \label{eq:load_cost_update}
\end{equation}
The fixed-state cost increment depends on signed velocity through $(P_4)_{34}$, with $K_{\mathrm{rem}}$ unchanged. The budget $B_t=\sum_{k\in\mathcal I_t}[\Delta V_k^L]_+$ sums positive increments, where $[x]_+=\max(x,0)$ and $\mathcal I_t$ contains updates after the reference and before reward evaluation. It measures changes in cost rather than mechanical work.

\subsection{Window Memory and Slack}
We use $N=10$, $\Delta=0.02\,\mathrm{s}$, $\kappa=0.5\,\mathrm{s}^{-1}$, and $\epsilon=0.02$, giving a maximum comparison interval of $N\Delta=0.2\,\mathrm{s}$ (200\,ms). The minimum slack in~\eqref{eq:soft_decay_violation} is
\begin{equation}
 \zeta_t=\left[
 V_t-e^{-\kappa h\Delta}V_{\mathrm{ref}}-B_t-\epsilon h/N
 \right]_+.
 \label{eq:minimal_decay_slack}
\end{equation}
Normalizing by $1+V_{\mathrm{ref}}$ gives the normalized decay violation $\delta_t$. A violation produces a finite reward penalty and does not trigger action projection or episode termination.

The reward memory stores the fixed reference, accumulated positive load-cost increments, and elapsed transitions. Its update follows three steps:
\begin{enumerate}
\item \emph{Within a window.} Keep the reference fixed, increment $h$, accumulate the load budget, and evaluate~\eqref{eq:soft_decay_violation} after each transition. The current cost uses the load that acted during that transition.
\item \emph{At a window boundary.} Evaluate the full-window comparison before renewing the reference and restarting the elapsed count and budget. The new reference includes the load update at its own state, so this update is excluded from the new $B_t$.
\item \emph{At initialization or replay continuation.} At episode startup, comparison begins with the first transition. Padded sensor frames do not advance the window. Replay, imagined rollouts, and the Critic inherit the stored memory, so an imagined start inside a window continues the same comparison.
\end{enumerate}

\subsection{Reward and Physical Regularization}
The training reward is $r_t=r_t^{\mathrm{sys}}+r_t^{\mathrm{phys}}$, where the system term combines current state quality with recent improvement:
\begin{equation}
 r_t^{\mathrm{sys}}=\Delta\left[
       \frac{6w_{\mathrm{sup},t}}{1+V_t}-H_+(\delta_t)\right],
 \label{eq:system_reward}
\end{equation}
Equivalently, the components in~\eqref{eq:learning_reward} are $r_t^{\mathrm{bal}}=6\Delta w_{\mathrm{sup},t}/(1+V_t)$ and $\ell_t^{\mathrm{dec}}=\Delta H_+(\delta_t)$, with $r_t^{\mathrm{sys}}=r_t^{\mathrm{bal}}-\ell_t^{\mathrm{dec}}$.
The smooth support weight $w_{\mathrm{sup},t}\in[0,1]$ in~\eqref{eq:system_reward} is based on foot--board proximity, board footprint, and base clearance. It weights the positive state benefit, while the decay penalty remains active independently of support. The one-sided Huber penalty is
\begin{equation}
 H_+(x)=\begin{cases}
       x^2/2,&0\leq x\leq1,\\
       x-1/2,&x>1.
       \end{cases}
 \label{eq:one_sided_huber}
\end{equation}
The auxiliary terms in $r_t^{\mathrm{phys}}$ cover foot-edge, clearance, and board--ground geometry, along with board acceleration, body posture and height. They also regularize action variation, soft joint and motor limits, motor effort, and bounded body acceleration, including its angular component. The body-uprightness penalty is $-0.5\Delta(1-R_{33})^2$ in $r_t^{\mathrm{phys}}$. Reward accumulates continuously, with no one-shot success bonus.

\subsection{Conditional Window-End Bound}
Consider a fixed policy on an uninterrupted, board-supported trajectory. Physical parameters remain fixed, and load-dependent zero-cost balance states are feasible under the contact and actuator constraints. We evaluate the bound at the reference-update boundaries of the reward windows.

Let $W_k$ denote costs after load updates at non-overlapping $N$-step endpoints, and let $\rho=e^{-\kappa N\Delta}$. The full-window violation $\delta_k$ is evaluated before the next endpoint's load update. Define $\Gamma_k$ as the sum of positive fixed-state load-cost increments after endpoint $k$ through endpoint $k+1$, including that final update. Substituting $\zeta_k=(1+W_k)\delta_k$ into~\eqref{eq:soft_decay_violation} and adding the endpoint load increment gives
\begin{equation}
 W_{k+1}\leq(\rho+\delta_k)W_k+\Gamma_k+\epsilon+\delta_k.
 \label{eq:window_affine_comparison}
\end{equation}
If the actual violations satisfy $0\leq\delta_k\leq\bar\delta<1-\rho$ and $0\leq\Gamma_k\leq\bar\Gamma$, this is a scalar contraction with factor at most $\rho+\bar\delta<1$. Iterating it yields
\begin{equation}
 \limsup_{k\to\infty}W_k\leq
 \frac{\bar\Gamma+\epsilon+\bar\delta}{1-\rho-\bar\delta}.
 \label{eq:conditional_cost_bound}
\end{equation}
For fixed loads with $\epsilon=\delta_k=0$, the comparison reduces to $W_k\leq\rho^kW_0$. Let $\lambda_4=\lambda_{\min}(P_4)>0$ and $E_0=\tfrac12Mv_0^2$. The decomposition gives
\begin{equation}
 \begin{aligned}
 \|\bar e\|^2&\leq V/\lambda_4,\\
 K_{\mathrm{rel}}/E_0
   &=(v_{\parallel}/v_0)^2+K_{\mathrm{rem}}/E_0\\
   &\leq\max\{1,\lambda_4^{-1}\}V.
 \end{aligned}
 \label{eq:full_motion_bound}
\end{equation}
At window endpoints, this comparison bounds board and load errors, signed relative velocity, and full relative kinetic energy under the stated violation and load-increment bounds. Zero cost requires zero board angle, board rate, static torque, and relative motion. Body uprightness is handled by the auxiliary reward.

Endpoint costs include contact transients. The continuous comparison in~\eqref{eq:ideal_lyapunov_decay} applies on smooth phases and extends across contact jumps when $V^+\leq V^-$.

\section{Proprioceptive Architecture and Supervision}
\label{app:feedback_implementation}

\begin{table}[!ht]
\caption{Network layer widths. MLP entries include input and output dimensions, whereas Conv1D entries give channel counts.}
\label{tab:network_configuration}
\centering\footnotesize
\begin{tabular}{@{}p{0.36\columnwidth}p{0.57\columnwidth}@{}}
\toprule
Module & Architecture\\
\midrule
Shared leg encoder & MLP $94\to128\to64$\\
Body branch & MLP $60\to64\to32$\\
Ordered fusion & MLP $288\to640\to224\to64$\\
Body Observer & MLP $420\to128\to64\to30$\\
Temporal encoder & Conv1D $42\to64\to64\to6$\\
Each phase head & Linear $128\to2$\\
Physical decoder & Conv1D $6\to64\to64\to13$\\
Shared Actor & MLP $98\to2048\to512$\newline $\to256\to12$\\
Privileged encoder & MLP $113\to2048\to1024$\newline $\to512\to256\to64$\\
World model & MLP $158\to1024\to1024$\newline $\to1024\to38$\\
Critic & MLP $195\to256\to128\to1$\\
\bottomrule
\end{tabular}
\end{table}

\subsection{Inputs and Sensor Histories}
Let $\bar o_t$ denote the sensor frame normalized by running statistics shared across branches and history windows. It has 42 coordinates, with block dimensions $12,12,3,3,12$ in the order defined in~\eqref{eq:proprioceptive_observation}. Clean joint/body targets use the same statistics to preserve relative signal levels and amplitudes.

The controller and all feedback branches update at 50\,Hz. The temporal encoder uses the full 128-frame history sampled at 20\,ms intervals, spanning a nominal 2.56\,s window with 2.54\,s between the oldest and newest samples. The short-history encoder and Body Observer use only the latest ten frames. At reset, the first frame fills the history so that control can start immediately. Reconstruction masks exclude this repeated padding.

\subsection{Short-History Context and Body Observer}
\subsubsection{Context Encoder}
Each leg contributes ten frames of three joint deviations, three joint velocities, and three previous-command components. Appending a four-value one-hot leg identity to these 90 values preserves leg identity in the shared encoder. The shared leg multilayer perceptron (MLP) and body branch use exponential linear unit (ELU) activations and the widths in Table~\ref{tab:network_configuration}. The body branch processes ten frames of gyro and gravity.

The four ordered leg embeddings and the body embedding concatenate to 288 values for ordered fusion. Its output parameterizes a 32-dimensional Gaussian mean and scale. With $\mathcal N_2(x)=x/\max(\|x\|_2,10^{-6})$, the context is
\begin{equation}
 \begin{aligned}
 \mu_t^S&=\mathcal N_2(\bar\mu_t^S),\\
 \sigma_t^S&=0.01+0.09\operatorname{sigmoid}(r_t^S),\\
 z_t^S(\varepsilon^z)&=\mathcal N_2(\mu_t^S+\sigma_t^S\odot\varepsilon^z).
 \end{aligned}
 \label{eq:student_context_sampling}
\end{equation}
Here $\bar\mu_t^S$ and $r_t^S$ are the two 32-coordinate network outputs. Environment interaction and default deployment use $\varepsilon^z=0$. Imagined training holds a replay-stored Gaussian sample fixed over each episode segment.

\subsubsection{Body Observer}
The Body Observer is a separate ELU MLP (Table~\ref{tab:network_configuration}). With $\bar x_t$ stacking the 30 current normalized joint and body coordinates, its estimate is $\widehat x_t=\bar x_t+O_\psi(\bar o_{t-9:t})$.
The observer is supervised on replay states collected by the deployment policy with the fixed-variance Gaussian reconstruction loss
\begin{equation}
 L_{\mathrm{obs}}=\frac1{2\cdot30}\sum_j
       \left(\frac{\widehat x_{t,j}-x_{t,j}^{\mathrm{clean}}}
                        {s_j^{\mathrm{sensor}}}\right)^2.
 \label{eq:body_observer_supervision}
\end{equation}
Here $s_j^{\mathrm{sensor}}$ is the sensor standard deviation in normalized units, obtained from the uniform noise half-width divided by $\sqrt3$.

\subsection{Temporal Encoding and Spectral Parameters}
The temporal encoder maps a $42\times128$ sensor history to six length-128 signals using the one-dimensional convolutional (Conv1D) layers in Table~\ref{tab:network_configuration}. Kernel size 9, stride 1, and padding 4 in each layer give a local receptive field of 25 frames. Batch normalization and ELU follow each convolution. Centered padding stays within the available history, whose latest sample is the current observation.

With $H_{\mathrm{hist}}=128$ and sampling interval $\Delta=0.02\,\mathrm{s}$, let $C_{ik}=\operatorname{rFFT}(y_i)_k$, $w_{ik}=|C_{ik}|^2$, and $\nu_k=k/(H_{\mathrm{hist}}\Delta)$. Following FLD's parameter extraction, frequency and amplitude use nonzero-frequency bins, while the offset comes from the zero-frequency (DC) coefficient:
\begin{equation}
 \begin{aligned}
 f_i&=\frac{\sum_{k>0}\nu_k w_{ik}}{\sum_{k>0}w_{ik}},\\
 A_i&=\frac{2}{H_{\mathrm{hist}}}\sqrt{\sum_{k>0}w_{ik}},\qquad
 b_i=\frac{\operatorname{Re}C_{i0}}{H_{\mathrm{hist}}}.
 \end{aligned}
 \label{eq:fourier_parameters}
\end{equation}
Spectral extraction uses all non-DC bins of the encoded signal. Numerical safeguards handle zero power and zero-amplitude derivatives while preserving zero amplitude.

Each channel has a phase head $\mathcal P_i$ consisting of Linear$(128,2)$ and batch normalization:
\begin{equation}
 \begin{aligned}
 (s_{xi},s_{yi})&=\mathcal P_i(y_i),\\
 \phi_i&=\operatorname{atan2}(s_{yi},s_{xi}).
 \end{aligned}
 \label{eq:learned_phase}
\end{equation}
Phase is measured in radians and learned through physical reconstruction and policy objectives. Each two-coordinate head defines one phase, giving six phases alongside 18 spectral scalars.

The feedback in~\eqref{eq:fourier_feedback_features} uses $f_0=4\,\mathrm{Hz}$ and enters the Actor separately from the spherical context in~\eqref{eq:student_context_sampling}. The frequencies and amplitudes describe learned signals, distinct from the physical kinetic energy defined in~\eqref{eq:relative_kinetic_energy}.

Frequency, amplitude, and offset remain fixed within a generated segment as phase advances with $\tau$. At the next control step, all parameters are recomputed from the sliding history.

\subsection{Physical Reconstruction and Continuation}
The extracted parameters generate latent trajectories on the grid
\begin{equation}
 \tau_j=j\Delta,\qquad j=-127,\ldots,10,\qquad
 \Delta=0.02\,\mathrm{s}.
 \label{eq:feedback_query_grid}
\end{equation}
The grid contains 138 points: 128 historical/current samples and ten continuation samples at 20--200\,ms. The decoder receives only latent trajectories synthesized from the current parameters on this grid. Temporal convolutions with kernel size 9, stride 1, and padding 4 preserve sequence length, using the channel widths in Table~\ref{tab:network_configuration}. Hidden layers use batch normalization and ELU, followed by a linear output.

The physical target is
\begin{equation}
 \begin{aligned}
 \chi_t=\big[&R_t^{\mathsf T}n_t^B,\;R_t^{\mathsf T}\omega_t^B,\\
             &R_t^{\mathsf T}\{v_t-\omega_t^B\times(p_t-p^H)\},\;
               R_t^{\mathsf T}\omega_t,\;\theta_t\big]\in\mathbb R^{13},
 \end{aligned}
 \label{eq:feedback_physical_targets}
\end{equation}
where $n_t^B$ is the board normal, $\omega_t^B=\dot\theta_t\mathbf e_y$, and $\theta_t$ is the scalar hinge angle relative to level. The four vector blocks are expressed in the robot frame. Predicted board normals are unit normalized. Smooth-L1 errors for the vector blocks use scales $(0.2,1\,\mathrm{rad/s},0.5\,\mathrm{m/s},1\,\mathrm{rad/s})$, repeated over the corresponding three coordinates. A positive scale $\theta_{\mathrm{sup}}$ normalizes the hinge-angle error.

The history loss includes all valid historical/current labels, and the future loss includes valid continuation labels. Each group is averaged over its valid coordinates and times. The representation objective is
\begin{equation}
 L_{\mathrm{repr}}=
 L_{\mathrm{history}}+
 \lambda_{\mathrm{future}}L_{\mathrm{future}}.
 \label{eq:representation_objective}
\end{equation}
The weight $\lambda_{\mathrm{future}}\geq0$ controls the contribution of short-term continuation. Current and historical labels come from replay collected by both roles, whereas continuation starts are collected by the deployment policy. Each loss excludes reset padding, replay cuts, invalid states, and truncation through validity masks.

Future labels are invalidated when the load schedule is activated, its target-generation seed changes, or its timing restarts. Valid historical labels from the same sample remain usable, and contact changes alone are retained.

The temporal encoder and phase heads produce motion feedback from proprioceptive history. The physical decoder sees only synthesized latent trajectories, without raw-sensor or Body Observer inputs, and its continuation targets follow recorded behavior. Action-conditioned policy rollouts use the separate world model. Sections~\ref{sec:multiscale_feedback} and~\ref{sec:privileged_learning} specify the deployment-policy inputs and privileged information.

\subsection{Actor and Deployment}
The Actor in~\eqref{eq:student_feedback_actor} has input dimension $32+30+12+24=98$ and ELU hidden layers with widths in Table~\ref{tab:network_configuration}. Both policies compute proprioceptive motion feedback from their respective histories. Each control step recomputes the complete temporal encoding, FFT parameters, and learned phase.

Deployment retains sensor normalization, the short-history encoder, Body Observer, temporal encoder, FFT, phase heads, amplitude-weighted coordinates, and shared Actor. The physical decoder and privileged training components are removed.

\section{Policy Training and Implementation}
\label{app:training_implementation}

\subsection{Privileged Information and Residual Context}
The privileged encoder uses ELU hidden activations and the architecture in Table~\ref{tab:network_configuration}. Its input comprises 42 current noisy sensor and previous-command values, two 17-value board-relative geometry frames, 30 clean body coordinates, three true body-frame base-velocity coordinates, and four foot--board contact labels. Geometry includes pivot displacement and board normal in the body frame, board-local base position, hinge angle and rate, board dimensions and limits, and heading descriptors. The clean body variables use the shared sensor normalization described in Appendix~\ref{app:feedback_implementation}.

The privileged encoder produces two 32-coordinate outputs: the raw residual $\bar d_t^T$ and a context scale. In~\eqref{eq:privileged_residual_mean}, $c_T=0.5/\sqrt{32}$ bounds the residual norm by 0.5 before normalization. The scale satisfies $\sigma_t^T=0.01+0.09\operatorname{sigmoid}(r_t^T)$, and context sampling follows the projected rule in~\eqref{eq:student_context_sampling}. The deployment context mean enters through residual addition, not as an input to the privileged encoder.

Physical simulation supplies contact labels for collected trajectories, while a learned current-state support readout supplies imagined contact inputs. Both roles derive motion feedback from their own sensor histories.

\subsection{Deployment-Policy-Derived Start Distribution}
Privileged closed-loop world-model rollouts use selected deployment-policy replay states as part of their start distribution. Table~\ref{tab:training_configuration} gives the environment allocations and imagined-start replacement rates. Replacements are divided equally between low-value and pre-failure states. All candidates must be valid and nonterminal, with finite value estimates. Low-value candidates are ranked within each recorded difficulty level to avoid cross-level comparisons.

Pre-failure sampling prioritizes distinct failure events within the interval and event cap in Table~\ref{tab:training_configuration}. A path to failure must contain no replay cut or invalid transition. Missing event samples fall back to valid deployment-policy candidates.

Each replacement preserves physical state, load context, sensor history, and finite-window reward memory. The remaining privileged starts and all deployment-policy starts use the ordinary replay sampler. These start allocations are independent of the equal role weights in~\eqref{eq:joint_policy_objective}.

\begin{table}[!ht]
\caption{Policy rollout and sampling configuration. Paired entries list the privileged policy first and the deployment policy second.}
\label{tab:training_configuration}
\centering\small
\begin{tabular}{@{}p{0.58\columnwidth}p{0.34\columnwidth}@{}}
\toprule
Setting & Value\\
\midrule
Privileged / deployment environments & 3072 / 1024\\
Privileged / deployment imagined starts & 3072 / 1024\\
Imagined trajectories per update & 4096\\
Rollout horizon $H$ & 32 steps (0.64 s)\\
Privileged-start replacement & 30\%\\
Privileged starts: low-value / pre-failure & 15\% / 15\%\\
Low-value candidate fraction & Bottom 20\% per level\\
Pre-failure interval & 5--25 steps (100--500 ms)\\
Maximum starts per failure & 4\\
World-model supervised horizon & 8 steps\\
Discount $\gamma$ & 0.99\\
Critic return parameter $\lambda$ & 0.95\\
Bootstrap / target bounds & $[-25,25]$\\
Privileged / deployment return weights & $1/2$, $1/2$\\
\bottomrule
\end{tabular}
\end{table}

\subsection{World Model and Critic}
The world model predicts residuals for structured integration. Its MLP (Table~\ref{tab:network_configuration}) receives invariant state features, commands, randomized dynamics, board geometry, loads, delay, and foot--board gaps. The 39-dimensional stored state contains position (3), quaternion (4), linear/angular velocities (6), and joint positions/velocities (26). The 13 joints comprise 12 actuated robot joints and the unactuated hinge. The distinct, 39-dimensional invariant input representation uses heading-removed, six-dimensional orientation features.

The 38 outputs comprise three-dimensional position, rotation, linear-velocity, and angular-velocity residuals, and 13-dimensional joint-position and joint-velocity residuals. The rotation vector, distinct from Euler angles, updates orientation through $R_{t+1}=R_t\exp([\delta\boldsymbol\theta_{\mathrm{eff}}]_\times)$, followed by quaternion storage in $wxyz$ order. Effective position, rotation, and joint-position increments combine learned residuals with velocity integration weighted 0.375 for old and 0.625 for new velocities. Environment-collected trajectories supervise eight-step rollouts. Reward and soft termination are evaluated from physical state. Delayed sensing uses endpoint interpolation and fresh noise.

A single Critic (Table~\ref{tab:network_configuration}) receives privileged physical and context features, geometry, role, previous command, contact or support information, and finite-window reward memory. Because it excludes motion feedback $m_t$ and full policy history, its value is a conditional approximation for the history-dependent controller. Training uses $\lambda$-returns with $\lambda=0.95$. Bootstrap values and target returns are bounded to $[-25,25]$, while the raw Critic output remains unconstrained.

\subsection{Differentiable Returns and Context Distillation}
Policy updates use the batch size and horizon in Table~\ref{tab:training_configuration}.

For role $r\in\{S,T\}$, the imagined return is
\begin{equation}
 \begin{aligned}
 \widehat J_r={}&\sum_{j=0}^{H-1}\gamma^j c_j r_j+
       \gamma^H c_H\overline U(\widehat s_H,\xi_H,r),\\
 c_j={}&\prod_{\ell<j}(1-d_\ell),\qquad\gamma=0.99,
 \end{aligned}
 \label{eq:imagined_role_return}
\end{equation}
where $d_\ell$ is analytic soft termination and $r_j$ includes both system and physical regularization rewards. $\overline U$ is a Critic bootstrap clipped to $[-25,25]$, distinct from the system cost $V$. Reward, survival weights, and closed-loop state transitions remain differentiable in the Actor objective.

Let $p_S$ and $q_T$ denote the deployment and privileged Gaussian context distributions before spherical projection. With $\operatorname{sg}$ denoting stop-gradient, their coordinate-mean distillation loss is
\begin{equation}
 \begin{aligned}
 &L_{\mathrm{KL}}=\mathbb E_{\mathcal D_S}\frac1{32}\sum_j\Bigg[\log\frac{\sigma_j^S}{\operatorname{sg}(\sigma_j^T)}-\frac12\\[2pt]
 &\qquad+\frac{\operatorname{sg}((\sigma_j^T)^2)+
       (\operatorname{sg}(\mu_j^T)-\mu_j^S)^2}
       {2(\sigma_j^S)^2}\Bigg],
 \end{aligned}
 \label{eq:preprojection_distillation}
\end{equation}
where $\mathcal D_S$ contains replay states collected by the deployment policy. The loss supervises the short-history Gaussian context. Gradients through the full privileged distribution are stopped, including its residual dependence on the deployment context mean.

\subsection{Gradient Updates and Calibration}
Deployment-policy returns update the shared Actor, short-history encoder, temporal encoder, and phase heads. Privileged-policy returns additionally update the privileged encoder and reach the deployment context mean through residual addition. Both return paths differentiate through spectral extraction and the phase heads.

Distillation updates the deployment context mean and scale. Physical reconstruction updates the temporal encoder, phase heads, and physical decoder. Body Observer parameters receive only reconstruction gradients, although return differentiation retains its input Jacobians.

World-model and Critic parameters are fixed during policy updates, retaining state/action derivatives for BPTT. Both models are trained separately through supervised updates.

The nonnegative coefficients $\beta_{\mathrm{aux}}$ and $\lambda_{\mathrm{repr}}$ in~\eqref{eq:joint_policy_objective} weight auxiliary supervision. Appendix~\ref{app:feedback_implementation} defines the observer and representation losses and validity masks.

Physical reconstruction uses detached replay inputs and labels, with optional replay warm-up. Batch-normalization statistics are calibrated on replay and fixed during joint training, imagined rollouts, and deployment. Convolutional weights, phase heads, and normalization affine parameters remain trainable.

\subsubsection{Representation Diagnostics}
Collected replay covers load compensation, residual oscillation, and quiet balance. Representation diagnostics include channel amplitudes, signal variance and correlation, frequency distributions, and blockwise physical reconstruction errors. Parameter shuffling probes reconstruction dependence, while masking or shuffling Actor features probes policy dependence.

\end{document}